\documentclass[conference]{IEEEtran}
\IEEEoverridecommandlockouts

\usepackage{cite}
\usepackage{float}
\usepackage{amsmath,amssymb,amsfonts}
\usepackage{bm}
\usepackage{amsthm}
\newtheorem{theorem}{Theorem}
\usepackage{algorithmic}
\usepackage{graphicx}
\usepackage{textcomp}
\usepackage{xcolor}
\DeclareMathOperator{\diag}{diag}

\def\BibTeX{{\rm B\kern-.05em{\sc i\kern-.025em b}\kern-.08em
    T\kern-.1667em\lower.7ex\hbox{E}\kern-.125emX}}

\newcommand{\norm}[1]{\lVert#1\rVert}
\newcommand{\R}{\mathbb{R}}
\DeclareMathOperator{\bdiag}{bdiag}

\begin{document}

\title{Energy-Based Physics-Informed Form Finding for Clustered Tensegrity Structures}

\author{Jing Qin$^{*}$, Muhao Chen$^{\dag}$
\thanks{$^{*}$Corresponding author. Email: jing.qin@uky.edu.}
\\
$^{*}$Department of Mathematics, University of Kentucky, Lexington, KY 40506, USA\\
$^{\dag}$Department of Mechanical and Aerospace Engineering, University of Houston, Houston, TX 77204, USA}

\maketitle

\begin{abstract}
Tensegrity form-finding and physical property prediction are fundamental problems in structural mechanics, which aim to determine equilibrium configurations and internal force distributions. These problems are challenging due to strong nonlinearity arising from the coupling between geometry and forces, and the need to satisfy equilibrium, stability, and structural constraints. This paper proposes an energy-based learning approach for clustered tensegrity form finding and physical property prediction. The proposed approach incorporates total potential energy minimization and constitutive relations into the training objective, enabling the prediction of equilibrium nodal configurations and the reconstruction of physical quantities such as member forces and force densities. By integrating energy-based physical losses directly into the learning process, the method promotes physical consistency while combining data-driven learning with physics-based constraints. Numerical experiments on tensegrity prism and lander structures demonstrate accurate prediction of equilibrium configurations and internal forces across different training-data ratios, indicating the potential of the proposed approach for nonlinear tensegrity form finding and structural analysis.
\end{abstract}


\section{Introduction}

Tensegrity structures achieve stable load-bearing behavior through self-equilibrated interactions between isolated compression components and a continuous tension network, enabling lightweight, deployable, and mechanically efficient structural designs. These properties make tensegrity structures attractive in applications such as aerospace engineering, robotics, and architectural design. A central challenge in the analysis and design of tensegrity systems is the form-finding problem, which aims to identify equilibrium configurations together with the corresponding internal force distributions that satisfy both static equilibrium and structural constraints \cite{chen2026self}.

From a mathematical perspective, determining a tensegrity equilibrium  configuration amounts to solving a nonlinear inverse problem in which structural geometry and internal force distribution must jointly satisfy equilibrium and constitutive constraints.
A broad spectrum of computational approaches has been developed to address this challenge. Among the earliest methods, the force density method (FDM) converts the equilibrium equations to a linear system by introducing member force densities, leading to fast computation but reduced flexibility for inverse analysis \cite{wang2021form}. Dynamic relaxation techniques subsequently introduced pseudo-dynamic iterative procedures for computing equilibrium states and were later extended to accommodate additional structural constraints, including inextensible members \cite{he2024modified}. Optimization-based strategies have also been investigated, such as least-squares formulations solved by the Levenberg--Marquardt algorithm \cite{yuan2017form} and inverse form-finding methods that cast equilibrium recovery as a parameter estimation problem \cite{dong2019inverse}. Recent studies have shown that exploiting the inherent low-rank structure of the force density matrix through optimization techniques can improve the efficiency and accuracy of tensegrity form finding \cite{wang2021form}.

The rapid growth of simulation and experimental datasets has promoted the adoption of machine learning techniques for tensegrity analysis. In particular, deep neural networks have been utilized to model the nonlinear mapping from structural variables to equilibrium configurations, providing alternatives to conventional numerical solvers \cite{lee2022deep}. This direction has been further extended to simultaneously estimate both geometric configurations and structural properties within a unified learning approach \cite{chen2024form}. To improve physical consistency, physics-informed neural networks (PINNs) enforce equilibrium equations and geometric constraints during network training, enabling data-free prediction of equilibrium solutions \cite{wang2025two}. Related variational and energy-based neural formulations employ energy principles as alternatives to pointwise residual minimization for mechanics problems \cite{nguyen2020deep,wang2024dcem}. Graph neural networks have also been introduced to exploit the connectivity patterns of tensegrity structures for scalable force and form prediction \cite{zhang2024graph}. Nevertheless, residual-based physics enforcement can introduce additional derivative evaluations and competing loss terms, potentially increasing training cost and optimization difficulty. Moreover, the clustered topology of tensegrity structures remains largely underexploited, motivating reduced-order physics-informed approaches.

In this work, we address these limitations by proposing an energy-based physics-informed learning approach for tensegrity form-finding that incorporates clustering and total potential energy minimization. Specifically, clustered tensegrity topology can reduce the number of independent member forces by representing multiple string segments within a clustered string through a common force. Rather than enforcing governing equations only through residual minimization, our method learns a parametric map from member rest-length variations to equilibrium nodal configurations by incorporating physics via energy-based loss functions derived from total potential energy minimization and constitutive relations. The resulting governing relations for nodal configurations and internal forces are combined with data-driven loss terms for nodal coordinates, member forces, and force densities. By embedding energy-based loss terms directly into the learning process, the proposed approach promotes physical consistency. Moreover, we apply a three-phase optimization strategy. Numerical experiments on tensegrity prism and lander structures demonstrate accurate prediction of equilibrium configurations and structural properties.

The remainder of this paper is organized as follows.  Section~\ref{sec:method} introduces the clustering-based reduction for member forces, clustered tensegrity equilibrium equations, and modal information. Then we describe the proposed form-finding method, which integrates data-driven, physics-informed, and energy-based losses with a three-phase optimization strategy. Section~\ref{sec:exp} presents numerical results for representative tensegrity structures, including the prism and the lander. Finally, Section~\ref{sec:con} concludes the paper and discusses future work.

\section{Proposed Method}\label{sec:method}

\subsection{Clustering-Based Reduction for Member Forces}\label{sec:cts}

Consider a tensegrity structure with $n_e$ elements, including bars and strings. A clustered tensegrity structure allows multiple adjacent string segments to form a single continuous string passing over intermediate joints via pulleys, such that the constituent segments share a common tension \cite{ma2022dynamics}.
Let $n_{ec}$ be the number of effective elements in a clustered tensegrity system, including bars, non-clustered strings, and clustered strings. The mapping from the non-clustered tensegrity structure to its clustered counterpart is represented by a clustering matrix $\bm Q\in\R^{n_{ec}\times n_e}$ whose $(i,j)$th entry is defined as
\begin{equation}\label{eqn:Q}
    q_{ij}=\left\{
\begin{aligned}
    &1,&&\mbox{if the $j$th element belongs to the $i$th }\\
    &&&\mbox{clustered element};\\
    &0,&&\mbox{otherwise}.
\end{aligned}
    \right.
\end{equation}
Then the reduced member force vector $\bm t_c\in\R^{n_{ec}}$ and the full counterpart $\bm t\in\R^{n_e}$ satisfy
\begin{equation}
    \bm t=\bm Q^T\bm t_c,
\end{equation}
where $\bm Q^T$ is the transpose of the matrix $\bm Q$. In particular,  conventional tensegrity can be viewed as a special case of clustered tensegrity when $\bm Q$ is the identity matrix.

\subsection{Clustered Tensegrity Equilibrium Equations}

Consider a tensegrity structure with $n_n$ nodes, consisting of $n_a$ free nodes and $n_b$ constrained nodes, where $n_n = n_a + n_b$. In addition, the structure has $n_e$ elements, consisting of $n_c$ bars and $n_s$ strings, where $n_c+n_s=n_e\geq n_{ec}$. The nodal coordinate matrix is denoted by
$
\bm{N}=[\bm{n}_1,\ldots,\bm{n}_{n_n}]\in\mathbb{R}^{3\times n_n}
$,
whose columns are the coordinate vectors of the nodes. The corresponding vectorized nodal coordinates at equilibrium are denoted by $\bm n\in\R^{3n_n}$, and the free nodal coordinate vector is denoted by $\bm n_a$. We use $\bm E_a\in\R^{3n_n\times 3n_a}$ to denote the index matrix that maps the free nodal indices to the full nodal indices such that $\bm n_a=\bm E_a^T\bm n$. Similarly, we denote the extraction matrix that maps the constrained nodal indices to the full nodal indices by $\bm E_b$. The full nodal coordinates can be recovered via
\begin{equation}\label{eqn:full_n}
    \bm n=\bm E_a\bm n_a+\bm E_b\bm n_b.
\end{equation}
Assume that $\bm C\in\R^{n_e\times n_n}$ is a connectivity matrix between tensegrity elements, including bars and strings, and nodes.

Let $\bm{x}$ and $\bm{l}$ denote the respective vectors of member force densities and member lengths. Then the equilibrium conditions of a clustered tensegrity structure are given by the following equations involving the nodal coordinate vector $\bm{n}\in\mathbb{R}^{3n_n}$ and the clustered member force vector
$\bm{t}_c\in\mathbb{R}^{n_{ec}}$ \cite{ma2022tensegrity}:
\begin{align}\label{eqn:Kn}
\bm{E}_{a}^{T}\bm{K}\bm{n} & =\bm{E}_{a}^{T}(\bm{f}_{e x}-\bm{g}),\\ \label{eqn:At}
\bm{E}_{a}^{T}\bm{A}_{t} \bm{t}_c & =\bm{E}_{a}^{T}(\bm{f}_{e x}-\bm{g}),
\end{align}
where the respective stiffness and the equilibrium matrices are defined as
\begin{align}
    \bm{K} &=(\bm{C}^T\diag(\bm{x})\bm{C})\otimes\bm{{ I}}_3,\\
    \bm{A}_1&=(\bm{C}^T\otimes \bm{I}_3)\bdiag(\bm{N}\bm{C}^T)\diag(\bm l)^{-1},\\
    \bm{A}_t& = \bm A_1\bm Q^T.
\end{align}
Here $\bdiag(\cdot)$ denotes the block-diagonalization operator that places the columns of a matrix along the block diagonal, while $\diag(\cdot)$ denotes the diagonalization operator that places the entries of a vector on the main diagonal.

Since the stiffness matrix $\bm{K}$ depends on the nodal coordinates $\bm{n}$, the equilibrium equation \eqref{eqn:Kn} is nonlinear in $\bm{n}$. Given the prescribed loading and initial structural configuration, the corresponding equilibrium configuration can be obtained using the procedure described in  \cite[Algorithm 1]{chen2024form}. From \eqref{eqn:Kn} and \eqref{eqn:At}, we can see that both $\bm n$ and $\bm t_c$ can be treated as functions of the prescribed rest length vector $\bm\ell=\bm\ell_0+\mathrm{d}\bm\ell$, where $\bm \ell_0$ and $\mathrm{d}\bm\ell$ denote the initial rest length vector and its change, respectively.

\begin{theorem}[Tensegrity Modal Information \cite{ma2022dynamics}]\label{modal}
The modal properties of a tensegrity structure, including its natural frequencies and vibration modes, can be obtained by solving the following generalized eigenvalue problem:
\begin{align}
& \bm{K}_{Taa} \bm{\varphi}=\omega^{2}\bm{M}_{aa} \bm{\varphi},
\label{eig_problem}
\end{align}
where
\[\begin{aligned}
& \bm{M} =\frac {1} {6}(|\bm{C}|^T\diag(\bm{m})|\bm{C}|+ \lfloor|\bm{C}|^T\diag(\bm{m})|\bm{C}|\rfloor)\otimes \bm{{I}}_3, \\
&    \bm{K}_{Taa}  = \bm{E}_a^T \bm{K}_T \bm{E}_a, ~ \bm{M}_{aa} = \bm{E}_a^T \bm{M} \bm{E}_a,\\
&    \bm{K}_T = \bm K +  \bm{A}_t \diag(\bm{e}_c)\diag(\bm{a}_c)\diag(\bm{l}_{0c})^{-1}  \bm{A}_t^T.\\
\end{aligned}\]
Here, $\bm{M}$ is the global mass matrix, $\bm m\in\mathbb{R}^{n_e}$ is the elemental mass vector, and the operator $\lfloor\cdot\rfloor$ sets the off-diagonal entries of a matrix to zero. The vectors $\bm{e}_c,\bm{a}_c,\bm l_{0c}\in\mathbb{R}^{n_{ec}}$ denote the Young's moduli, cross-sectional areas, and rest lengths of the clustered elements, respectively. Moreover, $\omega$ and $\bm{\varphi}$ denote the natural frequency and the corresponding mode shape, respectively.
\end{theorem}

\subsection{Proposed Method}

The proposed framework integrates supervised learning, variational energy minimization, and constitutive reconstruction within a unified learning architecture.
The network is trained to approximate a nonlinear operator that maps the prescribed member rest-length variations to the corresponding equilibrium configurations, from which the associated physical quantities are reconstructed. Specifically, given the initial rest length vector $\bm\ell_0$ and the rest-length change $\mathrm{d}\bm\ell$, a neural network learns the mapping
\begin{equation}\label{eqn:dl}
\mathrm{d}\bm\ell\mapsto \bm n,
\end{equation}
where $\bm n$ denotes the equilibrium nodal coordinates. Then both the clustered member force vector $\bm t_c$ and the member force density vector $\bm x$ can be explicitly expressed in terms of $\bm n$.

Unlike conventional PINNs that enforce governing-equation residuals pointwise, our energy-based model incorporates physics through energy-based loss functions derived from the total potential energy and constitutive relations. This formulation is motivated by the variational nature of tensegrity equilibrium, where stable equilibrium configurations are characterized as local minima of the total potential energy, modulo rigid-body motions when no displacement constraints are imposed. By embedding energy-based physical constraints directly into network training, the proposed approach promotes physically consistent equilibrium configurations.

Let $\mathcal{N}_\theta$ denote a neural network parameterized by $\theta$. Given the input rest-length change $\mathrm{d}\bm\ell\in\R^{n_{ec}}$ associated with clustered members, the network predicts the full equilibrium nodal-coordinate vector
\begin{equation}\label{eqn:n}
\widehat{\bm n}
=
\mathcal{N}_\theta(\mathrm{d} \bm \ell)\in\mathbb{R}^{3n_n}.
\end{equation}
Using the predicted nodal coordinates, the member length vector $\widehat{\bm l}$ and clustered member length vector $\widehat{\bm l}_c$ are computed from the tensegrity geometric relations
\[
\widehat{l}_i
=
\|(\widehat{\bm N}\bm C^T)_{:,i}\|_2,\quad i=1,\ldots,n_e,
\qquad
\widehat{\bm l}_c
=
\bm Q\widehat{\bm l},
\]
where $(\widehat{\bm N}\bm C^T)_{:,i}$ is the $i$th column of $\widehat{\bm N}\bm C^T$, and $\widehat{\bm N}\in\R^{3\times n_n}$ is the matrix form of $\widehat{\bm n}$. The clustered member forces are reconstructed through the constitutive relations
\begin{equation}\label{eqn:t_c}
\widehat{\bm t}_c
=
\diag(\bm e_c)\diag(\bm a_c)
\diag(\bm l_{0c})^{-1}
(\widehat{\bm l}_c-\bm l_{0c}),
\end{equation}
where $\bm l_{0c}$, $\bm e_c$, and $\bm a_c$ are defined in Theorem~\ref{modal}. Strings are tension-only, with $\hat t_{c,j}=0$ for negative strain, while bars follow \eqref{eqn:t_c}. The corresponding member force densities are constructed as
\begin{equation}\label{eqn:x}
\widehat{\bm x}
=
\diag(\widehat{\bm l})^{-1}\bm Q^T\widehat{\bm t}_c.
\end{equation}

To enforce physical consistency, we incorporate both equilibrium residual loss and energy-based physics loss into the training process. The equilibrium residual associated with the predicted configuration is defined as
\[
\widehat{\bm r}
=
\bm E_a^T
\left(
\widehat{\bm K}\widehat{\bm n}
-
(\bm f_{ex}-\bm g)
\right),
\]
where
\[
\widehat{\bm K}
=
(\bm C^T\diag(\widehat{\bm x})\bm C)\otimes \bm I_3.
\]
The corresponding residual loss over $m$ training samples is given by
\begin{equation}\label{eqn:Lr}
\mathcal{L}_{r}
=
\frac1m
\sum_{i=1}^m
\|
\widehat{\bm r}^{(i)}
\|_2^2.
\end{equation}

For active members, the elastic strain energy of the clustered tensegrity system is defined as
\begin{equation}\label{eqn:U}
U(\widehat{\bm n})
=
\frac12
(\widehat{\bm l}_c-\bm l_{0c})^T
\diag(\bm e_c)\diag(\bm a_c)
\diag(\bm l_{0c})^{-1}
(\widehat{\bm l}_c-\bm l_{0c}).
\end{equation}
For slack strings, the corresponding strain-energy contribution is set to zero. Then the total potential energy is given by
\begin{equation}\label{eqn:Pi}
\Pi(\widehat{\bm n})
=
U(\widehat{\bm n})
-
(\bm f_{ex}-\bm g)^T\widehat{\bm n}.
\end{equation}
For an unconstrained structure in the absence of external and gravitational loads, \(\Pi=U\) is invariant under rigid-body translations and rotations; hence, energy minimization characterizes the equilibrium configuration modulo rigid-body motions. Minimizing $\Pi$ drives the predicted configurations toward lower-potential-energy states, thereby promoting energetically stable equilibrium configurations.

The overall learning objective comprises both data-driven and physics-informed loss terms. Given $m$ training samples, the data-driven loss for the nodal coordinates reads as
\begin{equation}\label{eqn:Ld}
\mathcal{L}_{d}
=
\frac1m
\sum_{i=1}^m
\|
\widehat{\bm n}^{(i)}
-
\bm n^{(i)}
\|_2^2.
\end{equation}
To further improve the prediction of structural properties, we define the force prediction loss as
\begin{equation}\label{eqn:Lf}
\mathcal{L}_{f}
=
\frac1{m}
\sum_{i=1}^m
\bigg(\frac1{n_{ec}s_t^2}\|
\widehat{\bm t}_c^{(i)}
-
\bm t_c^{(i)}
\|_2^2+\frac1{n_es_x^2}\norm{\widehat{\bm x}^{(i)}-{\bm x}^{(i)}}_2^2\bigg).
\end{equation}
Here, $s_t$ and $s_x$ denote the root-mean-square (RMS) magnitudes of the reference member forces and force densities, respectively, computed from the training set for each train--test split. A lower bound of one is imposed on $s_t$ and $s_x$ to avoid amplification when the RMS magnitude is below one.
Based on \eqref{eqn:Pi}, we define the energy-based physics loss as
\begin{equation}\label{eqn:Le}
\mathcal{L}_{e}
=
\frac1m
\sum_{i=1}^m
\Pi(\widehat{\bm n}^{(i)}).
\end{equation}

The network parameters are optimized by minimizing a weighted sum of the data-driven and physics-informed losses defined in \eqref{eqn:Lr} and \eqref{eqn:Ld}--\eqref{eqn:Le}, i.e.,
\begin{equation}\label{eqn:L}
\mathcal{L}
=
\lambda_d\mathcal{L}_{d}
+
\lambda_f\mathcal{L}_{f}
+
\lambda_r\mathcal{L}_{r}
+
\lambda_e\mathcal{L}_{e},
\end{equation}
where $\lambda_d$, $\lambda_f$, $\lambda_r$, and $\lambda_e$ are weighting parameters selected empirically to balance the individual loss terms and adjusted progressively across the training phases. The data loss is emphasized throughout training, while the force and physics-informed losses are introduced sequentially.

Our learning architecture utilizes a multilayer perceptron to approximate the nonlinear mapping from the prescribed rest-length variations to the corresponding equilibrium tensegrity configurations. The network consists of multiple hidden layers with nonlinear activation functions. We employ the hyperbolic tangent activation function due to its smoothness and effectiveness for learning nonlinear structural responses. The output layer produces the full equilibrium nodal coordinate vector, from which the corresponding member lengths, clustered member forces, and force densities are reconstructed through the constitutive and geometric relations described previously.

To improve training stability and balance the competing objectives between data fitting and physics enforcement, we adopt a three-phase optimization procedure. The proposed staged training strategy gradually introduces force prediction and physics-informed constraints into the learning process.

In the first \textit{data-driven pretraining phase}, the neural network is trained using only the nodal coordinate prediction loss
\begin{equation}\label{eqn:L1}
\mathcal{L}^{(1)}
=
\lambda_d\mathcal{L}_{d}.
\end{equation}
This initialization stage enables the network to learn the coarse nonlinear mapping from rest-length changes to equilibrium configurations before introducing physics-informed constraints.

In the second \textit{force-assisted learning phase}, the force prediction loss is introduced
\begin{equation}\label{eqn:L2}
\mathcal{L}^{(2)}
=
\lambda_d\mathcal{L}_{d}
+
\lambda_f\mathcal{L}_{f}.
\end{equation}
At this stage, the network learns the equilibrium nodal coordinates under additional supervision from the reconstructed clustered member forces and force densities. The force supervision improves consistency between predicted geometries and internal structural responses.

In the final \textit{physics-informed optimization phase}, both the equilibrium residual loss and the energy-based physics loss are incorporated into the total objective
\begin{equation}\label{eqn:L3}
\mathcal{L}^{(3)}
=
\lambda_d\mathcal{L}_{d}
+
\lambda_f\mathcal{L}_{f}
+
\lambda_r\mathcal{L}_{r}
+
\lambda_e\mathcal{L}_{e}.
\end{equation}
The equilibrium-residual and energy losses promote equilibrium consistency and energetically stable configurations, respectively. For unconstrained systems, these conditions are understood modulo rigid-body motions.

For optimization, we first employ the Adam optimizer across all three phases. The network parameters are then refined using the limited-memory Broyden--Fletcher--Goldfarb--Shanno (L-BFGS) optimizer, which employs a quasi-Newton approximation to curvature information.

\section{Simulation Results}\label{sec:exp}
In this section, two representative tensegrity structures with different topological complexities, a prism and a lander, are selected to assess the proposed energy-based physics-informed approach for tensegrity form-finding and structural property prediction. The numerical examples use $\bm Q=\bm I$, while the formulation
generally accommodates clustered configurations through $\bm Q$. We report the
average relative error over 10 independent random train--test partitions. For a predicted
vector $\widehat{\mathbf{y}}$ and its reference vector $\mathbf{y}$,
the relative error (RE) for this prediction is defined as
\begin{equation}\label{eqn:re}
\mathrm{RE}(\widehat{\mathbf{y}},\mathbf{y})
=
{\|\widehat{\mathbf{y}}-\mathbf{y}\|_2}/
{\|\mathbf{y}\|_2}.
\end{equation}

Throughout the experiments, the datasets are generated using Algorithm~1 in~\cite{chen2024form}. External loads and gravity are neglected, such that $\bm f_{\rm ex}=\bm g=\bm 0$ and $\Pi=U$. All $3n_n$ nodal coordinates are free, so equilibrium is defined modulo rigid-body motions. A common numerical coordinate frame is maintained through a shared initial configuration and the supervised coordinate loss.

To investigate the influence of the amount of training data, we consider multiple train--test splitting ratios in the experiments. The datasets are randomly shuffled and partitioned into training and testing sets with training ratios selected from $\{0.7, 0.75, 0.8, 0.85, 0.9\}$. During training, all nodal coordinates and physical quantities are normalized using the training set statistics to improve numerical stability. The predicted coordinates are transformed back to physical units before evaluating the constitutive relations and physics-informed loss terms. The proposed approach learns equilibrium configurations from prescribed rest-length variations, while the associated physical quantities are reconstructed through constitutive relations and used in the physics-informed losses.

All experiments are conducted in Python using PyTorch on a desktop workstation with an Intel Core i9-9960X CPU, 64 GB RAM, and Windows 10 Pro. The proposed approach employs a fully connected neural network with five hidden layers, each containing 256 neurons, together with the hyperbolic tangent activation function by default. The network is trained using a three-phase optimization strategy with the Adam optimizer, followed by an L-BFGS refinement stage, as described in Section~\ref{sec:method}. The Adam optimizer is initialized with a learning rate of $10^{-3}$. The weighting parameters $(\lambda_d,\lambda_f,\lambda_r,\lambda_e)$ are set to $(1,0,0,0)$, $(20,1,0,0)$, and $(50,1,0.1,10^{-4})$ in the three Adam training phases, respectively, followed by $(100,1,0.1,10^{-4})$ during the L-BFGS refinement. For the training-ratio experiments, the three
Adam phases are trained for 300, 200, and 300 epochs, respectively, followed by at most 20 L-BFGS iterations.

Both bars and strings are modeled as steel (linear elastic) with a mass density of $7{,}850~\mathrm{kg/m^3}$, a Young's modulus of $200~\mathrm{GPa}$, and a yield strength of $300~\mathrm{MPa}$. The bars have hollow circular cross-sections with outer and inner radii of $10~\mathrm{mm}$ and $8~\mathrm{mm}$, respectively, whereas the strings are modeled as solid circular members with a radius of $2~\mathrm{mm}$. String forces are nonnegative, with zero force only in the unperturbed configuration.

\subsection{Experiment 1: Tensegrity Prism}

\begin{figure}[ht]
\centering
\includegraphics[width=.32\textwidth]{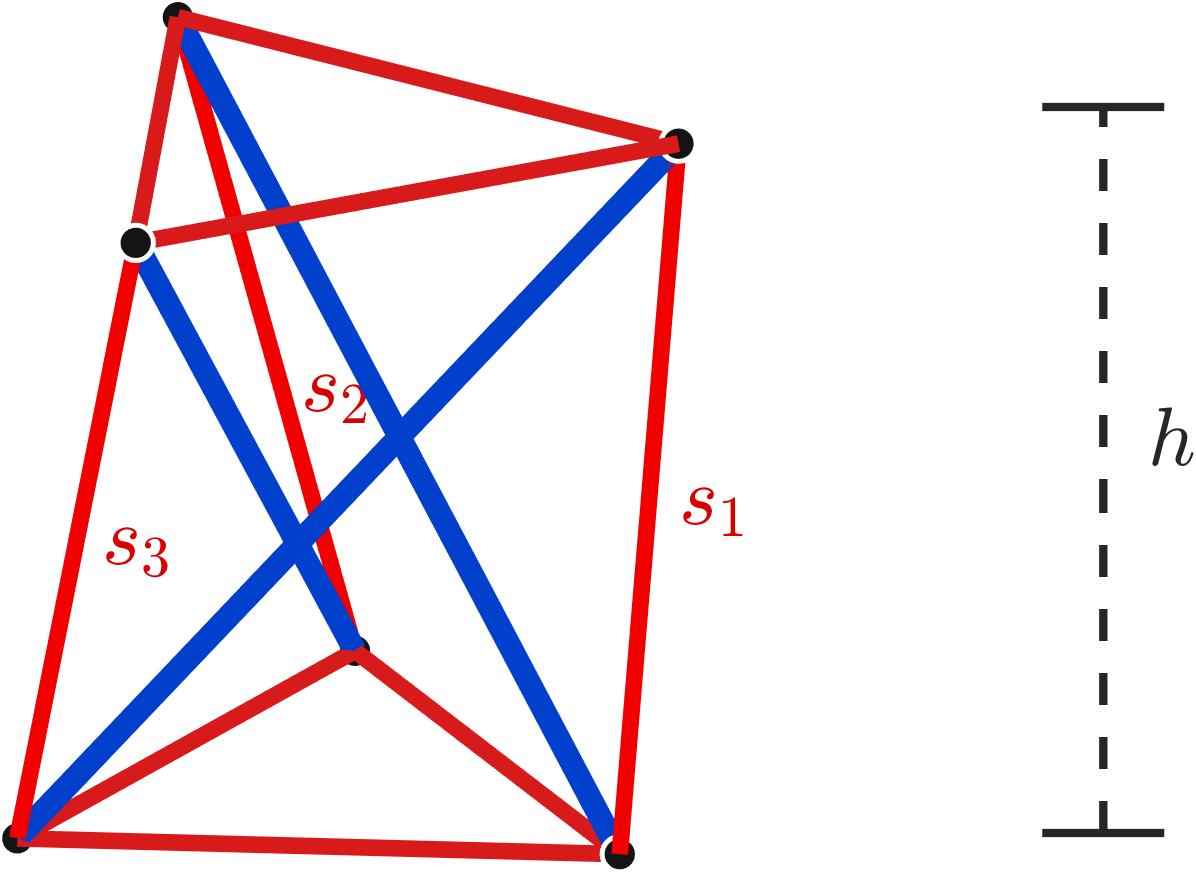}
\caption{Prism geometry with the three selected strings $s_1$, $s_2$, and $s_3$ used as rest-length change inputs. Strings are depicted in red and bars in blue. The prism has a height of $h=0.5$ m and a circumradius of $r=0.25$ m, defined by the circumscribed circle of the top-node triangle.
}\label{prism}
\end{figure}

\begin{figure}
    \centering    \includegraphics[width=.45\textwidth]{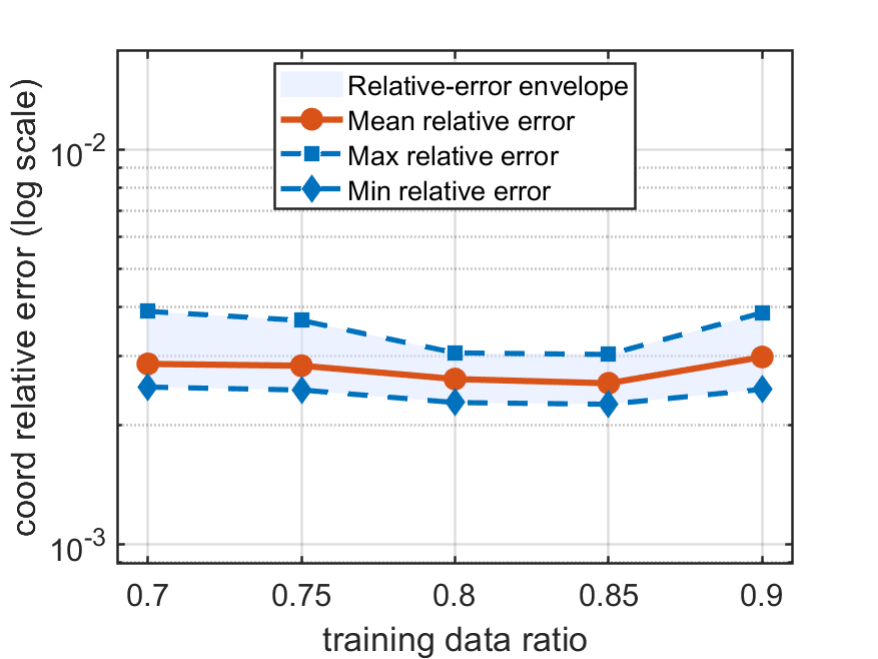}
     \vspace{-6pt}
    \caption{Relative errors of the predicted coordinates for the prism.}\label{fig:prism_coord_re}
    \vspace{-10pt}
\end{figure}

\begin{figure}
\vspace{-10pt}
    \centering    \includegraphics[width=.45\textwidth]{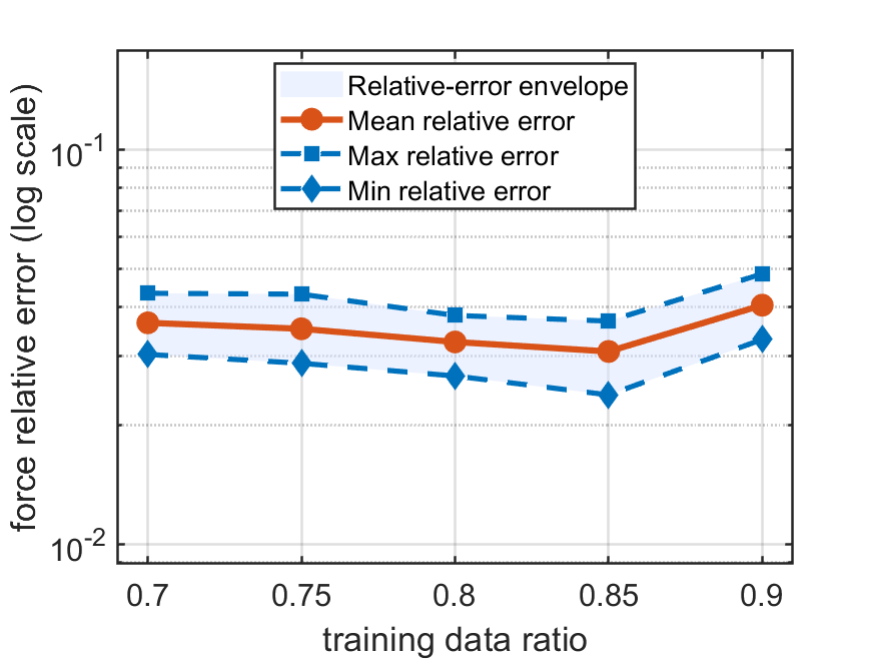}
     \vspace{-6pt}
    \caption{Relative errors of the predicted forces for the prism.}\label{fig:prism_force_re}
    \vspace{-10pt}
\end{figure}

We first evaluate the proposed method on the prism tensegrity structure shown in Fig.~\ref{prism}. A dataset of 1000 equilibrium configurations is generated using \cite[Algorithm~1]{chen2024form}. For each sample, the inputs are the rest-length variations of the three vertical strings, while the reference data consist of equilibrium nodal coordinates, member forces, and corresponding member force densities. Specifically, the rest-length variation of each selected string is sampled at 10 equally spaced values over $[-0.15,0]$ m, yielding 1000 equilibrium configurations. The proposed approach learns the nonlinear mapping from the prescribed rest-length variations to the equilibrium structural responses through the energy-based physics-informed losses and constitutive relations. The corresponding prediction errors under different training ratios are illustrated in Figs.~\ref{fig:prism_coord_re} and \ref{fig:prism_force_re}.

As shown in Fig.~\ref{fig:prism_coord_re}, our approach achieves accurate coordinate prediction for the prism over all tested training ratios. The coordinate relative error remains consistently on the order of $10^{-3}$, with limited variation among the independent trials, demonstrating reliable recovery of the equilibrium configuration even when only 70\% of the samples are used for training. As the training ratio increases, the coordinate prediction error generally decreases, indicating that the learned map benefits from additional equilibrium samples while maintaining consistent accuracy on the test set.

The force prediction results in Fig.~\ref{fig:prism_force_re} show a similar overall behavior. Compared with nodal coordinates, member forces are more sensitive to small geometric perturbations because they are determined by the constitutive relations and depend directly on member elongations. The force relative errors remain below approximately $10^{-1}$ across all train--test splits, indicating consistently accurate force prediction.  The reconstructed force densities achieve average relative errors ranging from approximately $3.3\times10^{-2}$ to $4.4\times10^{-2}$ across the considered training ratios. These results show that the proposed framework recovers not only the equilibrium shape but also physically consistent internal force states.

\subsection{Experiment 2: Tensegrity Lander}
\begin{figure}[ht]
\centering
\includegraphics[width=.235\textwidth]{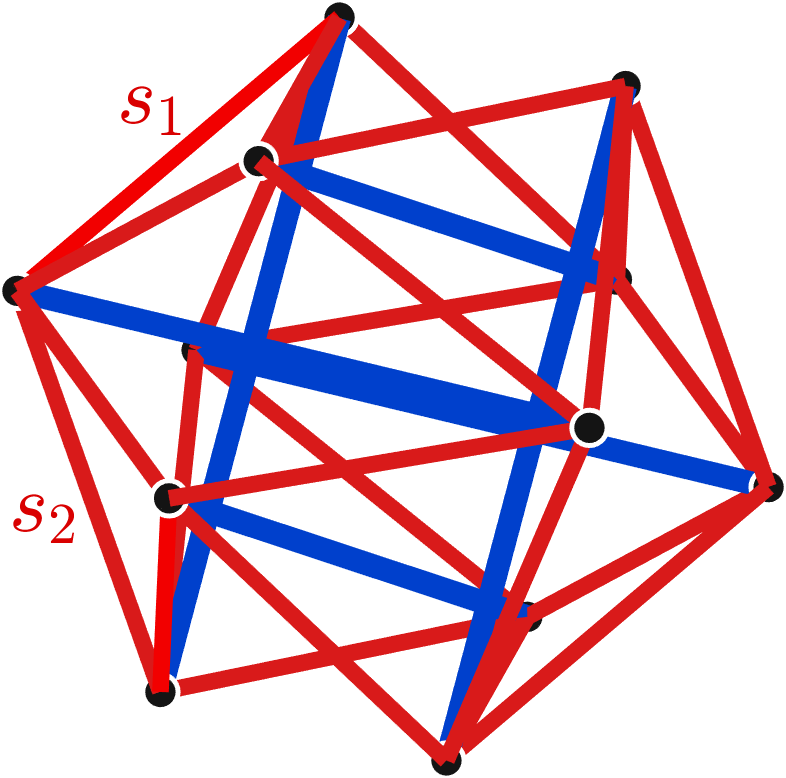}
\caption{Geometry of the six-bar tensegrity lander. The two strings $s_1$ and $s_2$ are selected as rest-length inputs. Strings are depicted in red and bars in blue. Each bar has a length of 1 m. }\label{fig:lander}
\end{figure}

The second experiment considers the tensegrity lander depicted in Fig. \ref{fig:lander}, comprising 6 bars and 24 strings.

Each data sample contains two cable rest-length variations as inputs and the corresponding equilibrium nodal coordinates, member forces, and force densities as reference data. The network directly predicts the 36 coordinate components of the 12 nodes, while the member forces and force densities are reconstructed from the predicted coordinates, prescribed rest lengths, and member lengths. Following the data generation procedure of the prism, the rest-length changes of two adjacent strings are each sampled at 10 equally spaced values over $[-0.3,0]$ m, yielding 100 equilibrium configurations.

\begin{figure}
    \centering    \includegraphics[width=.45\textwidth]{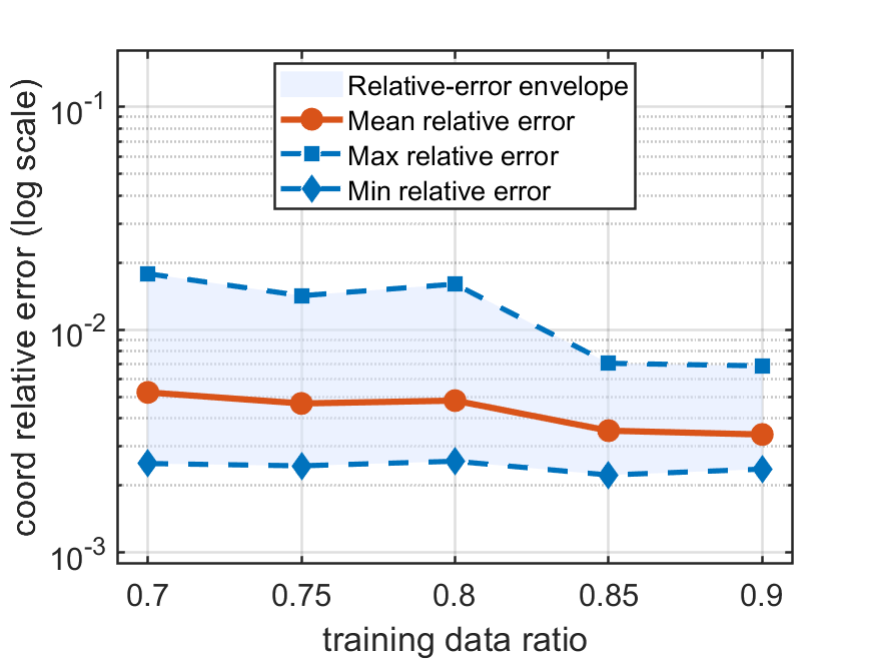}
     \vspace{-4pt}
    \caption{Relative errors of the predicted coordinates for the lander.}\label{fig:lander_coord_re}
\end{figure}

\begin{figure}
    \centering    \includegraphics[width=.45\textwidth]{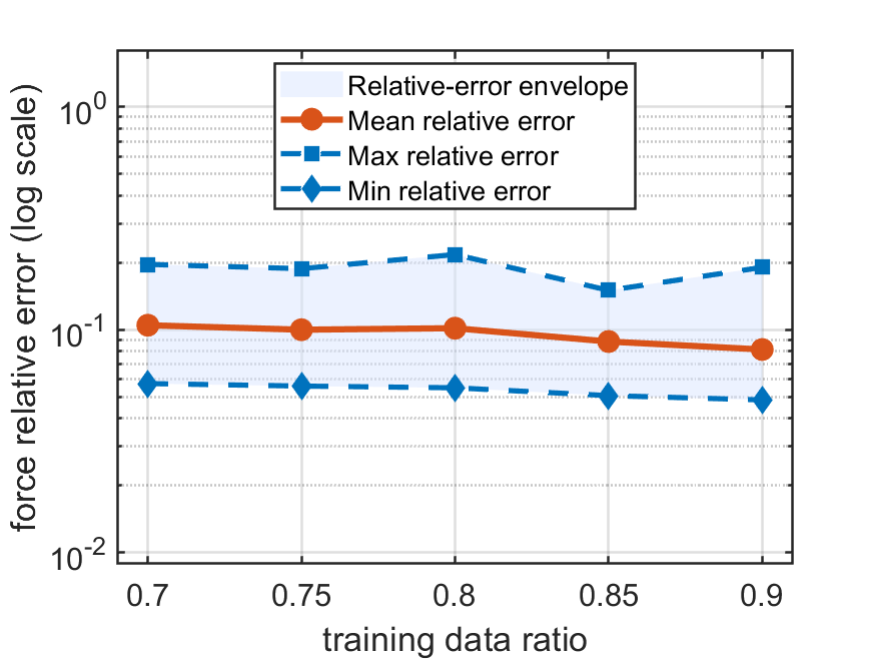}
    \vspace{-4pt}
    \caption{Relative errors of the predicted forces for the lander.}\label{fig:lander_force_re}
\end{figure}

The prediction errors under various training ratios are shown in Figs.~\ref{fig:lander_coord_re} and \ref{fig:lander_force_re}. Compared with the prism, the lander system has a higher-dimensional coordinate output and a more complex three-dimensional connectivity pattern, making the learning task more challenging. Even under this more complex setting, Fig.~\ref{fig:lander_coord_re} indicates that the proposed method produces accurate predictions of the equilibrium nodal coordinates for all considered training ratios. The reduction in error with increasing training data indicates that the network captures the nonlinear mapping between the two cable rest-length changes and the resulting equilibrium configuration. Despite its higher output dimension, the lander achieves comparable accuracy.

Fig.~\ref{fig:lander_force_re} further evaluates the prediction of member forces. The force errors generally decrease as the training ratio increases. The reconstructed force densities also achieve average relative errors ranging from approximately $7.1\times10^{-2}$ to $9.3\times10^{-2}$ across the considered training ratios. These results indicate that the predicted configurations are not only geometrically close to the reference solutions but also mechanically consistent with the corresponding internal force distributions. This is important for tensegrity form finding because accurate shape prediction alone does not guarantee a valid self-stress state. The lander experiment further demonstrates the applicability of the proposed approach to a higher-dimensional tensegrity structure while maintaining geometric and physical accuracy.

\section{Conclusion}\label{sec:con}
This paper introduces a variational learning approach for tensegrity form finding that integrates supervised learning with energy minimization and constitutive reconstruction. Specifically, the proposed approach approximates a nonlinear operator that maps prescribed cable rest-length changes to equilibrium nodal configurations, from which the corresponding member forces and force densities are reconstructed. Unlike conventional physics-informed neural networks that primarily enforce pointwise residuals of governing equations, the proposed approach incorporates physics via energy-based loss functions derived from total potential energy minimization and constitutive relations. The combined equilibrium-residual and variational energy losses promote structural equilibrium and physical consistency during training.

The clustering formulation provides a mechanism for reducing the number of independent member-force variables by representing multiple string segments within a clustered string through a common force. A three-phase optimization strategy is further developed to stabilize the training procedure by progressively introducing coordinate, force, and physics-informed loss terms. Numerical experiments on prism and lander tensegrity systems demonstrate that the proposed approach predicts equilibrium configurations and structural properties with high accuracy across different training-data ratios.

Future work includes experimental validation using physical tensegrity prototypes, the development of symmetry-aware learning and adaptive sampling strategies, and extensions to large-scale deployable tensegrity structures.

\bibliographystyle{IEEEtran}
\bibliography{ref}

\end{document}